\documentclass{article}

\usepackage[final]{neurips_2023}

\usepackage[utf8]{inputenc} 
\usepackage[T1]{fontenc}
\usepackage{hyperref}
\usepackage{url}
\usepackage{booktabs}       
\usepackage{amsmath,amssymb,amsthm,amsfonts}
\usepackage{graphicx}
\usepackage{enumitem}       
\usepackage{tikz}
\usepackage{pgfplots}
\usepackage[numbers]{natbib}
\usetikzlibrary{shapes,arrows,positioning,backgrounds}

\usepgfplotslibrary{groupplots}
\pgfplotsset{compat=1.18}

\usepackage{fancyhdr}
\usepackage{xcolor}

\newtheorem{remark}{Remark}

\title{A Stochastic Dynamical Theory of LLM Self-Adversariality:\\
Modeling Severity Drift as a Critical Process}

\author{%
  Jack David Carson \\
  Department of Physics \\
  Massachusetts Institute of Technology \\
  \texttt{jdcarson@mit.edu}
}

\begin{document}

\maketitle
\thispagestyle{plain} 

\begin{abstract}
This paper introduces a continuous-time \emph{stochastic dynamical} framework for understanding how large language models (LLMs) may \emph{self-amplify} latent biases or toxicity through their own chain-of-thought reasoning. The model posits an instantaneous ``severity'' variable $x(t) \in [0,1]$ evolving under a stochastic differential equation (SDE) with a drift term $\mu(x)$ and diffusion $\sigma(x)$. Crucially, such a process can be consistently analyzed via the Fokker--Planck approach if each incremental step behaves nearly Markovian in severity space. The analysis investigates \emph{critical phenomena}, showing that certain parameter regimes create phase transitions from subcritical (self-correcting) to supercritical (runaway severity). The paper derives stationary distributions, first-passage times to harmful thresholds, and scaling laws near critical points. Finally, it highlights implications for \emph{agents} and extended LLM reasoning models: in principle, these equations might serve as a basis for \emph{formal verification} of whether a model remains stable or propagates bias over repeated inferences.
\end{abstract}

\vspace{-2mm}
\section{Introduction}
\vspace{-1mm}

When a large language model (LLM) produces text, it conditions each new token on prior tokens, effectively referencing its own chain-of-thought (CoT). Such an iterative, self-referential mechanism can be beneficial---improving reasoning \citep{wei2022chain}---but it may also \emph{amplify} latent misalignment. Even in the absence of explicit adversarial prompts, once a partial bias or toxic statement appears, subsequent reasoning steps can elaborate and intensify that negativity. This phenomenon can be called \emph{self-adversarial} escalation. Recent empirical work by Shaikh et Al. \citep{shaikh2023secondthoughtletsthink} demonstrates this risk concretely: they found that zero-shot CoT reasoning significantly increases the likelihood of harmful or biased outputs across multiple sensitive domains, with the effect becoming more pronounced in larger models.

While stepwise or discrete Markovian toy models have offered an initial conceptual lens, this paper proposes a \textbf{continuous-time stochastic differential equation (SDE)} approach that captures:
\begin{itemize}[leftmargin=10pt, itemsep=0pt]
    \item A \emph{drift term} $\mu(x)$ that encodes deterministic escalation or correction of severity.
    \item A \emph{diffusion term} $\sigma(x)$ capturing the inherent randomness in LLM sampling.
    \item \textbf{Phase transitions} wherein a small parameter change can push the system from subcritical (stable near $x=0$) to supercritical (runaway severity near $x=1$).
\end{itemize}

\textbf{Why Fokker--Planck?} In practice, each short time increment $\Delta t$ might correspond to generating a small batch of tokens in the chain-of-thought. If the severity $x(t + \Delta t)$ depends only on (1) the current severity $x(t)$ and (2) a well-defined noise process (stemming from the LLM's sampling randomness), then the \emph{one-step} transition is approximately Markov in $x$. Such memoryless behavior is imperfect but can hold if the relevant context about bias or negativity can be compressed into the scalar severity variable. 
When the time- (or step-) spacing $\Delta t$ is small, the transitions can be recast into an SDE limit, making the \emph{Fokker--Planck equation} an apt tool for analyzing the probability flow in severity space.

\vspace{-1mm}

\begin{figure*}[t]
\centering
\resizebox{\textwidth}{!}{
\begin{tikzpicture}[
    node distance=1.5cm,
    input/.style={
        rectangle, 
        draw=black!70, 
        fill=gray!5, 
        rounded corners=3pt,
        text width=2.5cm,
        align=center, 
        minimum height=1cm,
        font=\small
    },
    process/.style={
        ellipse, 
        draw=blue!60, 
        fill=blue!5, 
        text width=2.3cm,
        align=center,
        font=\small
    },
    output/.style={
        rectangle, 
        draw=green!50!black, 
        fill=green!5, 
        rounded corners=3pt,
        text width=2.5cm,
        align=center, 
        minimum height=1cm,
        font=\small
    },
    danger/.style={
        rectangle, 
        draw=red!50!black, 
        fill=red!5, 
        rounded corners=3pt,
        text width=2.5cm,
        align=center, 
        minimum height=1cm,
        font=\small
    },
    arrow/.style={->, >=stealth, thick}
]

\node[input] (prompt) {Initial Prompt \\ \textit{(e.g., "Analyze group differences")}}; 

\node[process, below=of prompt] (cot) {\textbf{Chain-of-Thought}\\ \footnotesize Reasoning Process};
\draw[arrow] (prompt) -- (cot);

\node[process, left=3cm of cot, fill=blue!10] (subcritical) {
    \textbf{Subcritical Path} \\
    \footnotesize
    "Consider historical context... Policy solutions..."
};
\draw[arrow, blue!60] (cot) -- (subcritical) 
    node[midway, above, text=black, font=\footnotesize] {Self-correcting};

\node[process, right=3cm of cot, fill=red!10] (supercritical) {
    \textbf{Supercritical Path} \\
    \footnotesize
    "Inherent differences... Therefore..."
};
\draw[arrow, red!60] (cot) -- (supercritical)
    node[midway, above, text=black, font=\footnotesize] {Self-amplifying};

\node[output, below=of subcritical] (safe) {
    \textbf{Aligned Output} \\
    \footnotesize
    Balanced analysis with\\systemic perspective
};
\draw[arrow, green!50!black] (subcritical) -- (safe);

\node[danger, below=of supercritical] (toxic) {
    \textbf{Misaligned Output} \\
    \footnotesize
    Biased conclusions with\\harmful implications
};
\draw[arrow, red!50!black] (supercritical) -- (toxic);

\draw[dashed, red!70, thick] ([yshift=0cm]cot.south) -- ([yshift=-3cm]cot.south) 
    node[midway, right, text width=2cm, font=\footnotesize] {
        \textbf{Critical Threshold}\\
        Point of irreversible\\bias amplification
    };

\node[draw, fill=white, text width=3cm, align=left, 
      right=0.3cm of toxic, font=\footnotesize] (key) {
    \textbf{Key Dynamics:}\\[2pt]
    \textcolor{blue!60}{$\rightarrow$ Subcritical}: Bias correction\\
    \textcolor{red!60}{$\rightarrow$ Supercritical}: Bias amplification\\
    \textcolor{green!50!black}{$\square$ Safe}: Aligned reasoning\\
    \textcolor{red!50!black}{$\square$ Toxic}: Harmful outcomes
};

\begin{scope}[on background layer]
    \fill[blue!5] ([xshift=-4cm, yshift=1cm]subcritical.north) 
        rectangle ([xshift=-1.5cm, yshift=-1cm]safe.south);
    \fill[red!5] ([xshift=1.5cm, yshift=1cm]supercritical.north) 
        rectangle ([xshift=4cm, yshift=-1cm]toxic.south);
\end{scope}

\end{tikzpicture}
}
\caption{Conceptual diagram of self-amplifying bias in LLM chain-of-thought reasoning. Starting from a neutral prompt, the reasoning process can follow either a subcritical path (where biases are corrected) or a supercritical path (where biases amplify). The critical threshold marks where bias amplification becomes irreversible, leading to divergent outcomes in terms of alignment.}
\label{fig:concept}
\end{figure*}
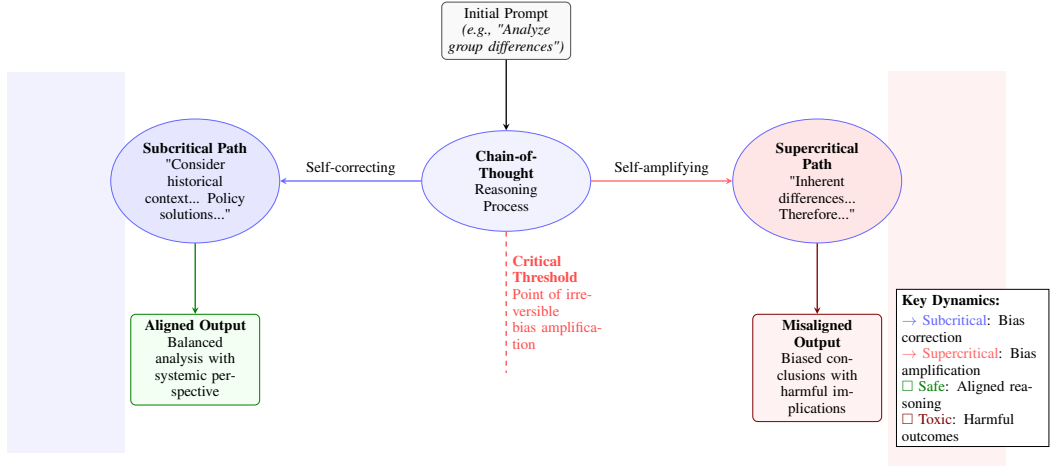
\section{Continuous-Time Severity Model}
\label{sec:sde_model}
\vspace{-1mm}

\subsection{State Variable and SDE}
Let $x(t)\in [0,1]$ represent the \emph{instantaneous severity} (e.g., toxicity, bias level) of the LLM's chain-of-thought at continuous time $t$. The simplest assumption is
\begin{align}\label{eq:sde_main}
dx(t) \;=\; \mu\bigl(x(t)\bigr)\,dt \;+\; \sigma\bigl(x(t)\bigr)\,dW(t),
\end{align}

\begin{itemize}[leftmargin=*]
    \item $\mu(x): [0,1]\to\mathbb{R}$ is the \emph{drift function}, capturing deterministic tendencies for severity to grow or diminish.
    \item $\sigma(x): [0,1]\to\mathbb{R}_{\ge 0}$ is the \emph{diffusion function}, capturing noise intensity at severity level $x$.
    \item $W(t)$ is a standard Wiener process (Brownian motion).
    \item We assume boundary conditions that keep $x(t)\in[0,1]$ (e.g., reflecting boundaries or saturating behavior).
\end{itemize}

\begin{remark}[Interpretation]
In practice, one might discretize  at time intervals $\Delta t$, obtaining $x_{t+\Delta t}-x_t \approx \mu(x_t)\Delta t + \sigma(x_t)\sqrt{\Delta t}\,\eta_t$, where $\eta_t\sim \mathcal{N}(0,1)$. This approximates the simpler Markov chain model but with \emph{continuous-state} transitions and explicit parametric forms of drift/diffusion.
\end{remark}

\subsection{Approximate Markov Assumption}
\label{sec:markov_argument}
\textbf{Why is $x(t)$ Markov?} If we measure severity at discrete intervals $\Delta t$, then $x(t+\Delta t)$ is not strictly memoryless, since the LLM can reference text from multiple earlier steps. However, if \emph{(a)} severity $x(t)$ effectively summarizes the "bias content" carried over from prior tokens, and \emph{(b)} the random generation of new tokens is conditionally independent (beyond severity), the system can be \emph{approximately} Markov at the level of $x(t)$. 

\paragraph{Inference for Fokker--Planck.}
In the limit $\Delta t \to 0$, standard diffusion-limit arguments \citep{gardiner2009stochastic} show that a well-defined drift $\mu(x)$ and diffusion $\sigma^2(x)$ yields a continuous-state Markov process. The corresponding \emph{Fokker--Planck} partial differential equation (PDE) then emerges naturally to describe the time evolution of the probability density $P(x,t)$ over severity. This, in essence, \emph{justifies} employing the continuum approach: if each short token block updates $x$ in a manner akin to a small Markov jump, we move to SDE as $\Delta t \to 0$.

\vspace{-1mm}
\section{Drift and Noise Terms}
\label{sec:drift_noise}
\vspace{-1mm}

\paragraph{Drift $\mu(x)$.}
The drift function is a phenomenological model capturing three key dynamics of severity evolution:
\[
\mu(x) = \underbrace{\alpha x(1-x)}_{\text{self-reinforcement}} - \underbrace{\beta x^2}_{\text{alignment}} + \underbrace{\gamma}_{\text{baseline}},
\]
where:
\begin{enumerate}[label=(\roman*), leftmargin=20pt]
    \item The logistic term $\alpha x(1-x)$ models self-reinforcing bias, borrowed from population dynamics, where severity grows but saturates as $x \to 1$. Parameter $\alpha$ reflects the model's tendency to elaborate on and amplify existing biases.
    
    \item The quadratic damping term $-\beta x^2$ represents alignment efforts (e.g., RLHF training) that counteract severity more strongly at higher $x$. Parameter $\beta$ quantifies the strength of bias suppression.
    
    \item The constant term $\gamma \geq 0$ captures spontaneous bias emergence from pretraining data or architecture, independent of prior reasoning steps.
\end{enumerate}

The equation admits closed-form solutions and exhibits critical phenomena analogous to phase transitions in physics.

\paragraph{Potential and Critical Behavior.}
The corresponding potential function $V(x)$ is obtained by integrating $-\mu(x)$. When $\alpha > \beta$, the drift remains positive above a critical threshold $x_c = \frac{\alpha - \beta}{\alpha + \beta}$, leading to supercritical (runaway) behavior. Conversely, when $\beta > \alpha$, the system remains subcritical, with severity returning to low values.

\paragraph{Supercritical vs. Subcritical.}
If $\alpha$ dominates $\beta$ in some range, the drift remains positive, causing $x(t)$ to \emph{increase} on average. Conversely, large $\beta$ ensures $x(t)$ gets pulled back to a stable equilibrium near 0. The first scenario is called "supercritical" and the second "subcritical."

\paragraph{Diffusion $\sigma(x)$.}
We can let $\sigma(x) = \sigma_0 + \sigma_1\,x$ (with $\sigma_0,\sigma_1\ge 0$), so that the process becomes more volatile at higher severity. This choice reflects the intuition that \emph{controversial or negative} lines of reasoning produce more varied or explosive expansions from the LLM.

\section{Fokker--Planck Equation and Stationary Behavior}
\label{sec:fpe}
The Fokker--Planck (FP) equation \citep{risken1996fpe} for the probability density $P(x,t)$ of $x(t)$ is:
\begin{equation}\label{eq:fp_equation}
\frac{\partial P}{\partial t} 
= -\frac{\partial}{\partial x}\Bigl[\mu(x)\,P\Bigr]
+\frac{1}{2}\,\frac{\partial^2}{\partial x^2}\Bigl[\sigma^2(x)\,P\Bigr].
\end{equation}
\textbf{Intuitive meaning}: 
\begin{itemize}[leftmargin=12pt,itemsep=1pt]
    \item The term $-\tfrac{\partial}{\partial x}[\mu(x)P]$ captures the \emph{deterministic flow} in severity space. 
    \item The second term $\tfrac12\tfrac{\partial^2}{\partial x^2}[\sigma^2(x)P]$ captures \emph{random spreading}.
\end{itemize}
\subsection{When is Fokker--Planck Valid?}
Because $x(t)$ is a \emph{(near) Markov process in continuous state}, it satisfies an SDE of the form \eqref{eq:sde_main}. The associated \emph{Kolmogorov forward equation} is precisely \eqref{eq:fp_equation}, known as the Fokker--Planck equation in the physics literature \citep{gardiner2009stochastic}. Hence, if we accept the Markov approximation from Section~\ref{sec:markov_argument}, the FP approach is the correct PDE for describing $P(x,t)$.

\subsection{Stationary Distribution \texorpdfstring{$P_{\mathrm{ss}}(x)$}{P\_ss(x)}}
When $\frac{\partial P}{\partial t}=0$, \eqref{eq:fp_equation} yields a \emph{stationary distribution} $P_{\mathrm{ss}}(x)$ satisfying
\begin{equation}\label{eq:Pss}
0 = -\frac{\partial}{\partial x}\Bigl[\mu(x)\,P_{\mathrm{ss}}(x)\Bigr]
+\frac{1}{2}\,\frac{\partial^2}{\partial x^2}\Bigl[\sigma^2(x)\,P_{\mathrm{ss}}(x)\Bigr].
\end{equation}
Standard results give the closed-form expression \citep{gardiner2009stochastic}
\[
P_{\mathrm{ss}}(x) \;\propto\;\frac{1}{\sigma^2(x)} 
\exp\Bigl(2\int_0^x \frac{\mu(z)}{\sigma^2(z)}\,dz\Bigr).
\]
If $\mu$ strongly favors growth near $x>0$, $P_{\mathrm{ss}}(x)$ may concentrate away from zero, or even become \emph{bimodal}. This is typically the sign of a \emph{supercritical} regime, where high-severity states are stable attractors.

\section{Critical Phenomena and Phase Transitions}
\label{sec:critical}

\subsection{Qualitative Picture of Criticality}

Consider the drift parameters $(\alpha,\beta,\gamma)$ from the logistic-like example:
\[
\mu(x)=\alpha x (1-x) - \beta x^2 + \gamma.
\]
\begin{enumerate}[leftmargin=15pt,itemsep=2pt]
    \item \textbf{If $\alpha<\beta$}, there is a stable fixed point near $x=0$. Severity remains small, with the noise occasionally pushing it upward but the drift pulling it back. This regime is called \emph{subcritical or aligned}.
    \item \textbf{If $\alpha>\beta$}, the drift can remain positive after $x$ surpasses some threshold, pushing it toward $x\approx 1$. This regime is called \emph{supercritical or runaway}. 
\end{enumerate}
Near the boundary $\alpha=\beta$, the system may exhibit \emph{critical slowing down} and increased fluctuations. In the Fokker--Planck landscape, $P_{\mathrm{ss}}$ can transition from unimodal (peaked at low $x$) to bimodal or peaked at high $x$.
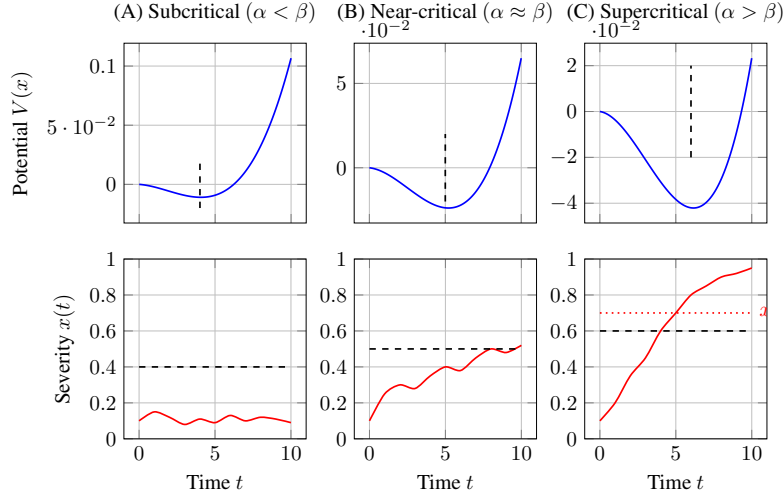
\begin{figure*}[t]
\centering
\begin{tikzpicture}[scale=0.8]
\begin{groupplot}[
    group style={
        group size=3 by 2,
        vertical sep=0.6cm,
        horizontal sep=0.8cm
    },
    width=0.33\textwidth,
    height=0.2\textheight,
    grid=major,
    every axis plot/.append style={thick},
]

\nextgroupplot[
    title={(A) Subcritical $(\alpha < \beta)$},
    ylabel={Potential $V(x)$},
    xticklabels=\empty
]
\addplot[blue, domain=0:1, samples=100] {-0.15*x^2 + 0.2667*x^3 - 0.01*x};
\addplot[black, dashed] coordinates {(0.4,-0.02) (0.4,0.02)};

\nextgroupplot[
    title={(B) Near-critical $(\alpha \approx \beta)$},
    xticklabels=\empty
]
\addplot[blue, domain=0:1, samples=100] {-0.225*x^2 + 0.3*x^3 - 0.01*x};
\addplot[black, dashed] coordinates {(0.5,-0.02) (0.5,0.02)};

\nextgroupplot[
    title={(C) Supercritical $(\alpha > \beta)$},
    xticklabels=\empty
]
\addplot[blue, domain=0:1, samples=100] {-0.3*x^2 + 0.3333*x^3 - 0.01*x};
\addplot[black, dashed] coordinates {(0.6,-0.02) (0.6,0.02)};

\nextgroupplot[
    ylabel={Severity $x(t)$},
    xlabel={Time $t$},
    ymin=0, ymax=1
]
\addplot[red, smooth] coordinates {
    (0,0.1) (1,0.15) (2,0.12) (3,0.08) (4,0.11) 
    (5,0.09) (6,0.13) (7,0.10) (8,0.12) (9,0.11) (10,0.09)
};
\addplot[black, dashed] coordinates {(0,0.4) (10,0.4)};

\nextgroupplot[
    xlabel={Time $t$},
    ymin=0, ymax=1
]
\addplot[red, smooth] coordinates {
    (0,0.1) (1,0.25) (2,0.3) (3,0.28) (4,0.35) 
    (5,0.4) (6,0.38) (7,0.45) (8,0.5) (9,0.48) (10,0.52)
};
\addplot[black, dashed] coordinates {(0,0.5) (10,0.5)};

\nextgroupplot[
    xlabel={Time $t$},
    ymin=0, ymax=1
]
\addplot[red, smooth] coordinates {
    (0,0.1) (1,0.2) (2,0.35) (3,0.45) (4,0.6) 
    (5,0.7) (6,0.8) (7,0.85) (8,0.9) (9,0.92) (10,0.95)
};
\addplot[black, dashed] coordinates {(0,0.6) (10,0.6)};
\addplot[red, dotted, thick] coordinates {(0,0.7) (10,0.7)}
node[right] {$x_{\text{harm}}$};

\end{groupplot}
\end{tikzpicture}
\caption{Self-amplifying bias dynamics in LLMs. Top row shows potential landscapes $V(x) = -\frac{\alpha}{2}x^2 + \frac{\alpha + \beta}{3}x^3 - \gamma x$ for different parameter regimes. Bottom row shows corresponding stochastic trajectories solving $dx(t) = \mu(x)dt + \sigma(x)dW(t)$, with drift $\mu(x) = \alpha x(1-x) - \beta x^2 + \gamma$ and noise $\sigma(x) = \sigma_0 + \sigma_1 x$. Dashed lines indicate critical thresholds $x_c$, and dotted line shows harmful threshold $x_{\text{harm}}$. Parameters: $\gamma = 0.01$, $\sigma_0 = 0.05$, $\sigma_1 = 0.1$.}
\label{fig:dynamics}
\end{figure*}
\subsection{Scaling Laws}
We expect, from parallels with nonequilibrium phase transitions \citep{odor2004universality}, that the correlation length $\xi$ and relaxation time $\tau$ might diverge near criticality. Formally:
\[
\xi(\Delta)\sim|\Delta|^{-\nu}, \quad
\tau(\Delta)\sim|\Delta|^{-z\nu}, 
\]
where $\Delta=\alpha-\beta$ measures how far the system is from the critical point. The exponents $(\nu,z)$ could, in principle, be measured by analyzing fluctuations of $x(t)$ in simulations or from real LLM logs.

\section{First-Passage Analysis of Harmful States}
\label{sec:first_passage}

A threshold $x_{\mathrm{harm}}\in (0,1)$ can be defined as the boundary beyond which the LLM's outputs are deemed severely toxic or misaligned. The \emph{first-passage time} $T$ is:
\[
T = \inf\{\,t\ge 0 : x(t)\ge x_{\mathrm{harm}}\}.
\]
One can derive a partial differential equation (PDE) for $\langle T\rangle(x)$, the expected time to blow-up starting from $x(0)=x$ \citep{gardiner2009stochastic}. In one dimension with SDE \eqref{eq:sde_main}, the boundary condition is $\langle T\rangle(x_{\mathrm{harm}})=0$, and one may impose $\tfrac{d\langle T\rangle}{dx}\big|_{x=0}=0$ if $x=0$ is reflecting. The solution generally shows an exponential sensitivity to integrals of $\mu/\sigma^2$, a hallmark of how quickly a supercritical drift can push $x(t)$ up to the harmful region.

\section{Implications for Agents and Extended Reasoning}
\label{sec:implications}

Beyond static text generation, modern LLMs can act as \emph{agents}, performing multi-step reasoning or planning over extended time horizons. In such scenarios, the \emph{severity} $x(t)$ can keep feeding back into the agent's policy or chain-of-thought. If, for example, the system is in a supercritical domain of parameters, we risk \emph{cascading bias} that leads to undesirable or disallowed outputs as the agent self-references prior negative statements.

\paragraph{Formal Verification Potential.}
A promising direction is using these SDE and Fokker--Planck equations for a \emph{rigorous check} of whether, under all typical sampling dynamics, the severity distribution remains \emph{stationary} near 0 or if it flows inevitably to high $x$. If we can bound $\mu(x)$ below $x$, or show that $P_{\mathrm{ss}}(x)$ is unimodal at low severity, this might serve as a formal proof of \emph{subcritical alignment} for an LLM-based agent. Conversely, detecting that $x_{\mathrm{harm}}$ is almost certainly reached within finite time would be a red flag, signifying \emph{runaway misalignment} in extended inference.

\paragraph{Interpretation and Safety Gains.}
In practice, evaluating $\mu(x)$ and $\sigma(x)$ from real LLM data would require carefully controlled experiments and robust severity metrics. But should the fit reveal that the system is "near critical," design teams might reduce $\alpha$ (the self-amplification) or increase $\beta$ (alignment damping) to ensure stable performance over many reasoning steps.

\section{Conclusion and Outlook}
\label{sec:conclusion}

This paper has presented a \textbf{stochastic differential equation framework} for modeling LLM chain-of-thought severity. By positing that severity $x(t)$ evolves in a near-Markov manner, the analysis shows how the \emph{Fokker--Planck equation} naturally arises to describe the probability flow in severity space. Crucially, small changes in drift parameters can yield a \textit{phase transition} from subcritical (safe or self-correcting) to supercritical (runaway) regimes. The work analyzes the stationary distribution, first-passage times to harmful thresholds, and near-critical scaling laws reminiscent of classical nonequilibrium physics.

\textbf{Implications for extended reasoning models} are profound: in principle, these equations open the door for \emph{formal verification} of stability or guaranteed subcritical behavior. Coupled with improved severity metrics and data-fitting procedures, the approach could help LLM developers ensure that multi-step, agentic reasoning systems do not inadvertently \emph{self-escalate} into severely misaligned outputs. Future work includes multi-dimensional expansions, memory kernels for more realistic references to older tokens, and bridging to interpretability methods that track which internal components of an LLM drive $\mu(x)$ at each stage. 

\begin{ack}
This paper would not be possible if not for the generous support of the MIT Schwartzman College of Computing and the MIT Social and Ethical Responsibilities of Computing (SERC) Fellowship. Funding was provided by SERC Research Fund. This research specifically was influenced by Dr. Amir Reisizadeh's SERC group on LLM debiasing.
\end{ack}

\small
\bibliographystyle{plainnat}

\end{document}